%% file: main.tex
\documentclass[sigconf]{acmart}
\AtBeginDocument{%
}

\copyrightyear{2025}
\acmYear{2025}
\setcopyright{acmlicensed}
\acmConference[MM '25] {Proceedings of the 33rd ACM International Conference on Multimedia}{October 27--31, 2025}{Dublin, Ireland.}
\acmBooktitle{Proceedings of the 33rd ACM International Conference on Multimedia (MM '25), October 27--31, 2025, Dublin, Ireland}
\acmISBN{979-8-4007-2035-2/2025/10}
\acmDOI{10.1145/3746027.3754722}

\input{math_commands}

\usepackage{overpic}
\usepackage{subcaption}
\usepackage{multirow}
\usepackage{multicol}

\usepackage{xcolor}
\PassOptionsToPackage{numbers}{natbib}
\definecolor{url}{RGB}{0,73,147}
\definecolor{mypink}{HTML}{bc4749}
\definecolor{mygray}{gray}{0.85}
\definecolor{rgbgray}{gray}{0.95}
\definecolor{MSMblue}{HTML}{DAE8FC}
\definecolor{Mgreen}{HTML}{D5E8D4}
\definecolor{MTMyellow}{HTML}{FFF2CC}

\newcommand{\xmark}{\ding{55}}%

\usepackage{array}
\usepackage{colortbl}

\usepackage{pifont}  % For \ding command
\usepackage{bbding}    % \Checkmark,\XSolid,... (需要和pifont宏包共同使用)

\usepackage{caption}
\usepackage{enumitem}
\usepackage{multirow}
\makeatletter
\newcommand{\thickhline}{%
\noalign {\ifnum 0=`}\fi \hrule height 1pt
\futurelet \reserved@a \@xhline
}
\makeatother
\newcommand{\pub}[1]{\color{gray}{\tiny{[{#1}]}}}

\newcommand{\bestr}[1]{\textbf{#1}}
\newcommand{\ebestr}[1]{#1}

\usepackage{xspace}

\makeatletter
\DeclareRobustCommand\onedot{\futurelet\@let@token\@onedot}
\def\@onedot{\ifx\@let@token.\else.\null\fi\xspace}

\def\eg{\emph{e.g}\onedot} 
\def\ie{\emph{i.e}\onedot}

\def\etal{\emph{et al}\onedot}
\makeatother

\usepackage{hyperref}

\begin{document}

\title{Motion Matters: Motion-guided Modulation Network for Skeleton-based Micro-Action Recognition}


\author{Jihao Gu}
\orcid{0009-0009-0141-4807}
\affiliation{
\institution{
University College London
}
\city{London}
\country{United Kingdom}
}
\email{jihao.gu.23@ucl.ac.uk }

\author{Kun Li}
\orcid{0000-0001-5083-2145}
\authornote{Corresponding author}
\affiliation{
\institution{
ReLER, CCAI, Zhejiang University
}
\city{Hangzhou}
\country{China}
}
\email{kunli.hfut@gmail.com}

\author{Fei Wang}
\orcid{0009-0004-1142-6434}
\affiliation{
\institution{
Hefei University of Technology
}
\institution{Institute of Artificial Intelligence, Hefei Comprehensive National Science Center}
\city{Hefei}
\country{China}
}
\email{feiwang@xinsight.com.cn}

\author{Yanyan Wei}
\orcid{0000-0001-8818-6740}
\affiliation{
\institution{
Hefei University of Technology
}
\institution{Intelligent Interconnected Systems Laboratory of Anhui Province}
\city{Hefei}
\country{China}
}
\email{weiyy@hfut.edu.cn}

\author{Zhiliang Wu}
\orcid{0000-0002-6597-8048}
\affiliation{
\institution{
ReLER, CCAI, Zhejiang University
}
\city{Hangzhou}
\country{China}
}
\email{wu_zhiliang@zju.edu.cn}

\author{Hehe Fan}
\orcid{0000-0001-9572-2345}
\affiliation{
\institution{
ReLER, CCAI, Zhejiang University
}
\city{Hangzhou}
\country{China}
}
\email{hehefan@zju.edu.cn}

\author{Meng Wang}
\orcid{0000-0002-3094-7735}
\affiliation{
\institution{Hefei University of Technology}
\city{Hefei}
\country{China}
}
\email{eric.mengwang@gmail.com}

\renewcommand{\shortauthors}{Jihao Gu et al.}

\begin{abstract}
Micro-Actions (MAs) are an important form of non-verbal communication in social interactions, with potential applications in human emotional analysis. However, existing methods in Micro-Action Recognition often overlook the inherent subtle changes in MAs, which limits the accuracy of distinguishing MAs with subtle changes. 
To address this issue, we present a novel \textbf{M}otion-guided \textbf{M}odulation \textbf{N}etwork (\textbf{MMN}) that implicitly captures and modulates subtle motion cues to enhance spatial-temporal representation learning. 
Specifically, we introduce a Motion-guided Skeletal Modulation module (MSM) to inject motion cues at the skeletal level, acting as a control signal to guide spatial representation modeling. In parallel, we design a Motion-guided Temporal Modulation module (MTM) to incorporate motion information at the frame level, facilitating the modeling of holistic motion patterns in micro-actions. 
Finally, we propose a motion consistency learning strategy to aggregate the motion cues from multi-scale features for micro-action classification. 
Experimental results on the Micro-Action 52 and iMiGUE datasets demonstrate that MMN achieves state-of-the-art performance in skeleton-based micro-action recognition, underscoring the importance of explicitly modeling subtle motion cues. The code will be available at \href{https://github.com/momiji-bit/MMN}{https://github.com/momiji-bit/MMN}.
\end{abstract}




\maketitle

\section{Introduction}
Micro-Action~\cite{liu2021imigue,chen2023smg,guo2024benchmarking,li2025prototypical,li2024mmad,liu2024micro} is a nonverbal language in social interaction~\cite{aviezer2012body}, has attracted extensive attention in recent years. In contrast to conventional human actions~\cite{soomro2012ucf101,kuehne2011hmdb,li2023vigt,li2023datae,li2025repetitive,wang2025exploiting}, micro-actions are characterized by imperceptible and low-intensity movements. These movements hold significant potential for understanding individuals' internal emotional states~\cite{li2024eald,zhao2025temporal}. 

In human-centric micro-action analysis~\cite{liu2021imigue,chen2023smg,balazia2022bodily,guo2024benchmarking,li2023joint,li2023data}, remarkable progress has been made. From the perspective of the dataset, Liu~\etal and Chen~\etal collected the iMiGUE~\cite{liu2021imigue} and SMG~\cite{chen2023smg} datasets that contain spontaneous micro-gestures in the upper limbs. These videos are sampled from the post-competition review of athletes, revealing deep emotional states conveyed through these micro-gestures. 
Meanwhile, Balazia~\etal~\cite{balazia2022bodily} presented the MPIIGI dataset, which annotated the subtle upper-body behaviors in group interactions and highlighted the social behaviors in human communications. 
To explore the whole body micro-actions, Guo~\etal~\cite{guo2024benchmarking} developed the large-scale Micro-Action 52 (MA-52) dataset comprising 20k video samples interviewed from 205 participants, distributed in 52 micro-action classes, benchmarking micro-action recognition. 
From the perspective of methodology, traditional action recognition methods, including 2D CNN based~\cite{wang2018temporal,lin2019tsm,feichtenhofer2019slowfast}, 3D CNN based~\cite{C3D,I3D}, GCN-based~\cite{yan2018spatial,liu2020disentangling}, and Transformer-based~\cite{liu2022video}, are selected to evaluate these datasets. However, these methods show suboptimal performance due to the challenges posed by the subtle and low-intensity nature of micro-actions. 
Recently, Li~\etal~\cite{li2025prototypical} proposed a novel prototypical ambiguous calibrating net to address the ambiguity in micro-actions. 

\begin{figure}[!t]
\centering
\includegraphics[width=1.0\linewidth]{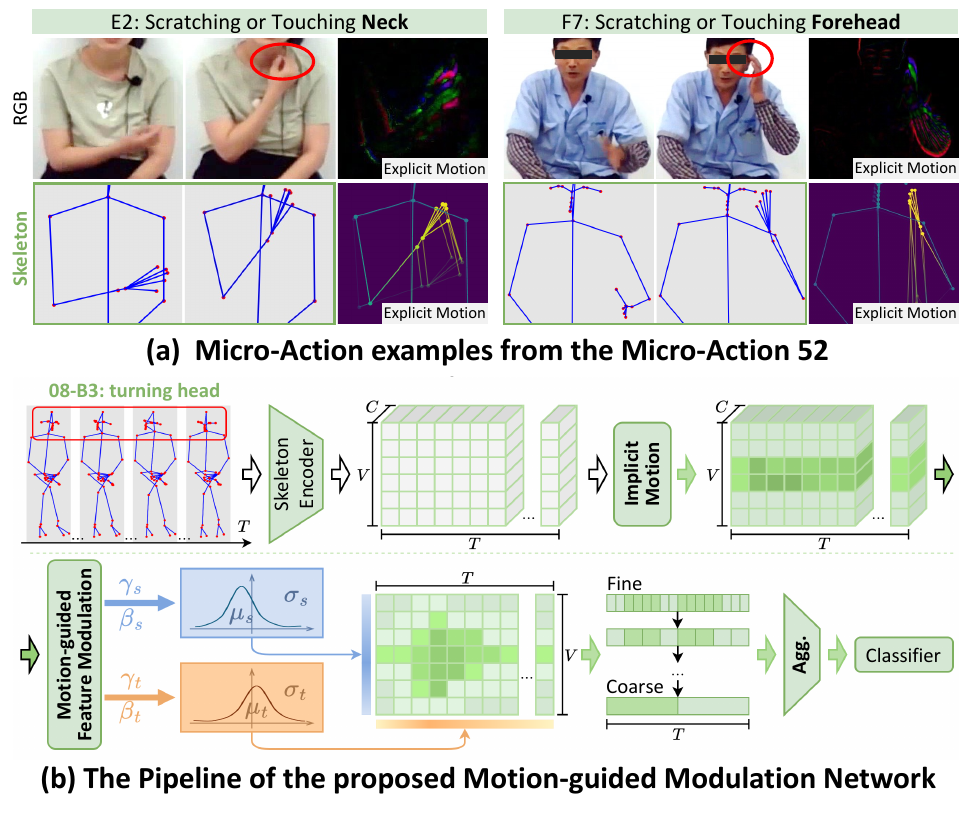}
\caption{(a) Micro-Action Recognition (MAR) aims to recognize micro-action with subtle motion amplitudes, high inter-class similarity, and notable intra-class variability caused by individual differences. 
(b) The Pipeline of the proposed Motion-guided Modulation Network (MMN). We attempt to dynamically modulate the motion information into skeletal and temporal separately, fascinating spatial-temporal skeleton feature representation.}
\label{fig:intro}
\end{figure}

Despite the above significance, accurately recognizing micro-actions remains particularly challenging due to several inherent difficulties. 
We depicted several micro-action examples in Figure~\ref{fig:intro} (a).
\textbf{1) Subtle motion amplitudes.} Micro-actions inherently exhibit low-amplitude movements, making them hard to distinguish from background movements. 
\textbf{2) High inter-class similarity.} Micro-actions usually involve only minor motion differences, leading to visually ambiguous and overlapping representations. For example, ``scratching or touching shoulder'' and ``scratching or touching neck'' are only different in the touching region. 
\textbf{3) Notable intra-class variability.} 
Due to individual differences in body shape, appearance, and other physical traits, the same micro-action will manifest in noticeably different ways across subjects.

Intuitively, capturing subtle temporal motions is key to effective micro-action recognition. Compared to the RGB modality, skeleton sequences provide a compact yet expressive representation that is inherently robust to appearance noise and well-suited for modeling fine-grained body motion~\cite{yan2018spatial,chen2021channel,duan2022dg,wang2024eulermormer,wang2024frequency}. 
Motivated by this, we attempt to model subtle temporal motions within skeleton sequences, enabling the network to effectively capture and emphasize subtle yet discriminative motion patterns. 
To this end, we propose a novel Motion-guided Modulation Network (MMN), specifically designed to amplify subtle motion cues critical for distinguishing micro-actions. As shown in Figure~\ref{fig:intro} (b), we decouple motion patterns into skeletal and temporal branches, facilitating more structured spatial-temporal skeleton feature representation. At the skeletal level, key joints are adaptively enhanced, while at the temporal level, informative frames are emphasized to capture meaningful motion patterns. 

The overview of the proposed Motion-guided Modulation Network (MMN) is illustrated in Figure~\ref{fig:method}. 
Firstly, we encode the input skeleton sequences using skeleton-aware embeddings into a high-dimensional feature space. 
Then, the proposed motion-guided feature modulation injects motion into skeletal and temporal modalities through a unified architecture to build motion-guided spatiotemporal features. 
Subsequently, the motion consistency learning module is designed to capture motion cues from fine-to-coarse spatiotemporal features. 
Finally, the learned features are used for micro-action classification. 

In summary, the contributions of this paper are summarized as follows:
\begin{itemize}
\item We conduct a comprehensive evaluation of existing skeleton-based action recognition methods in the context of micro-action recognition and establish standardized protocols for benchmark datasets. 
\item We present a motion-guided modulation network explicitly to capture subtle motion patterns and guide the learning of discriminative spatiotemporal features for micro-action recognition. 
\item Extensive experiments conducted on the Micro-Action 52 and iMiGUE datasets validate the effectiveness of the proposed method. Qualitative analysis highlights the effectiveness of the motion-guided modulation network in micro-action recognition. 
\end{itemize}

%%%%%%%%%%%%%%%%%%%%%%%%%%%%%%%%%%%%%
\section{Related Work}~\label{sec:related}

\subsection{Micro-Action Recognition}
Micro-Action Recognition (MAR)~\cite{liu2021imigue,chen2023smg,balazia2022bodily,guo2024benchmarking,li2025prototypical,chen2024prototype,guo2024mac,li2024advancing,gu2025mm,liu2025online} has emerged as a significant and challenging task within human-centric action analysis, aiming to classify subtle and low-intensity body movements. Recent advances in the field have been primarily driven by the release of large-scale benchmark datasets and the design of sophisticated model architectures. 
The iMiGUE dataset~\cite{liu2021imigue} collected identity-free, high-quality videos from 72 athletes during sports press conferences, thus enabling integrated analyses of both action and affective states.
These videos are annotated with 32 distinct micro-gesture categories. The SMG dataset~\cite{chen2023smg} presented a diverse collection of data supporting both micro-gesture recognition and detection. This dataset consists of recordings from 40 participants who narrated fabricated or real stories, during which subtle gestures and corresponding emotional states were recorded. 
However, these datasets are mainly focused on the upper limb micro-actions. To exploit full-body micro-actions, the MA-52 dataset~\cite{guo2024benchmarking} collected a large number of samples (22k) across 52 action-level and 7 body-level micro-action categories from professional psychological interviews. Thereby facilitating micro-action recognition of subtle visual differences.

Due to the inherent difficulty of distinguishing micro-actions characterized by subtle visual differences and high inter-class similarity, micro-action recognition (MAR) requires models to capture both spatial and temporal dynamics. Specifically, MANet ~\cite{guo2024benchmarking} designed a joint optimization strategy combining cross-entropy and embedding losses, aligning visual features more closely with action semantics and thereby improving the discrimination of visually similar micro-actions. 
To tackle the ambiguity arising from subtle inter-class visual differences, PCAN~\cite{li2025prototypical} introduced a hierarchical action-tree to identify easily confusable samples and employed a contrastive calibration module to refine the spatio-temporal feature. 

\subsection{Skeleton-based Action Recognition}
Skeleton-based action recognition~\cite{li2018independently,liu2016spatio,liu2017skeleton,perez2021interaction,duan2022pyskl,duan2022revisiting,do2024skateformer} modeled human body dynamics from both spatial and temporal perspectives using skeletal data. Existing approaches can be broadly categorized into the following groups: 
\textbf{RNN/CNN-based approaches}: Early research focused on using recurrent neural networks to handle the sequential nature of skeleton data~\cite{li2018independently,liu2016spatio,liu2017skeleton,perez2021interaction,si2019attention,zhang2017view,zhu2016co}. 
\textbf{GCN-based approaches}: Given that skeleton data naturally forms a graph structure, where joints are nodes and bones are edges. Graph Convolutional Networks have gained widespread attention~\cite{chen2021channel,chen2021multi,cheng2020decoupling,duan2022pyskl,korban2020ddgcn,kwon2021h2o,li2019actional,yan2018spatial}. 
\textbf{Transformer-based approaches}: Transformer-based architectures were introduced to skeleton-based action recognition for their ability to model long-range dependencies~\cite{qiu2023spatio,wang20233mformer}. 
\textbf{Motion-based approaches}: Motion differentials, such as temporal velocities or spatial bone displacements, have been widely used to enhance skeleton-based action recognition. Many methods explicitly compute first- or higher-order differences from raw skeletons as additional inputs or pseudo-images to highlight dynamic patterns~\cite{yang2023action,shi2019two,cheng2020skeleton,wang2024taylor}. However, explicit differencing may disproportionately amplify sensor or annotation noise, thus limiting the effectiveness of downstream models in distinguishing subtle actions.

While the aforementioned methods have achieved remarkable progress, they exhibit limitations in MAR primarily due to the inherent challenge of capturing subtle skeletal and temporal dynamics. Differing from previous works that utilized static, pre-computed motion differences, we propose a Motion-guided Modulation Network (MMN) specifically designed for MAR. Our approach integrates differential motion cues adaptively within the learned feature representations, dynamically enhancing model sensitivity to subtle micro-action variations.

\begin{figure*}[t]
\centering
\includegraphics[width=0.95\linewidth]{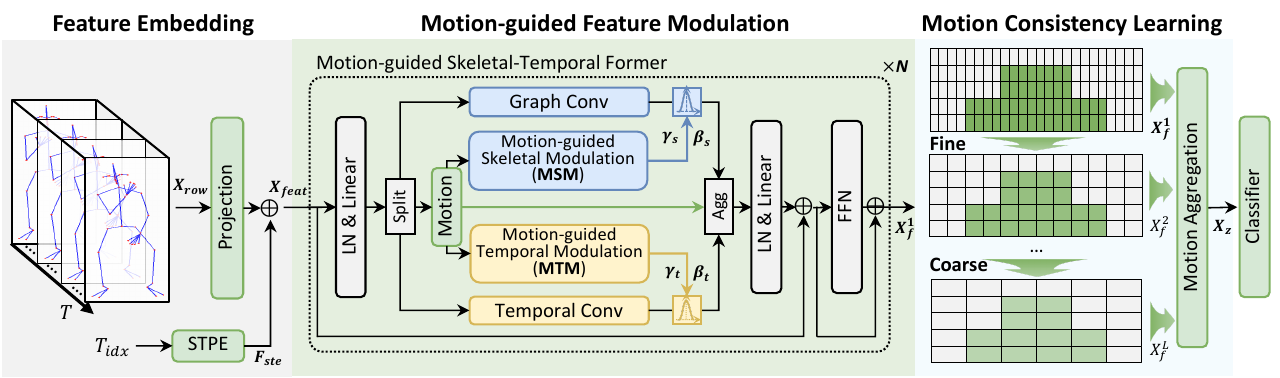}
\caption{Overview of the proposed Motion-guided Modulation Network (MMN).
It mainly consists of three key modules: Feature Embedding (\S~\ref{sec:embedding}), Motion-guided Feature Modulation (\S~\ref{sec:mfb}), and Motion Consistency Learning (\S~\ref{sec:mcl}). 
We first project the skeleton data into a high-dimensional feature space $\mX_{proj}$. Then, the motion-guided feature modulation module is designed to inject motion patterns into the spatial and temporal dimensions separately, guiding the model to mine crucial cues of micro-actions. Finally, the motion consistency learning module aggregates multi-scale motion cues for action classification. 
}
\label{fig:method}
\end{figure*}

%%%%%%%%%%%%%%%%%%%%%%%%%%%%%%%%%%%%%
\section{Methodology}~\label{sec:method}

\subsection{Problem Formulation} 
Micro-Action Recognition (MAR)~\cite{guo2024benchmarking,li2025prototypical} is a fundamental task in the field of micro-action analysis~\cite{liu2021imigue,chen2023smg,balazia2022bodily}. Previous works usually adopt RGB modality data to model the motion patterns. However, skeleton data is more compact and better at capturing subtle motion changes that are essential for micro-actions. In contrast, this paper focuses on Skeleton-based MAR, where only skeleton data is used as input. The task can be formulated as follows:
\begin{equation}
Y = \gM (\mX, \Theta),
\end{equation}
where $\mX=\{\mX_1, \mX_2, \mX_3, \ldots, \mX_T\}\in\R^{T\times V\times C_{in}}$ is the input skeleton, and $Y$ denotes the micro-action categories. $\gM$ is the optimized model parameters by $\Theta$.

\subsection{Network Overview}
As shown in Figure~\ref{fig:method}, the proposed Motion-guided Modulation Network (MMN) consists of three key modules: \S~\ref{sec:embedding} Feature Embedding module (FE), \S~\ref{sec:mfb} Motion-guided Feature Modulation module (MFM), and \S~\ref{sec:mcl} Motion Consistency Learning module (MCL). 
First, in the Feature Embedding module, we embed the input skeleton data with skeleton-aware embedding into a high-dimensional feature space. 
Then, in the MFM module, we propose motion-guided skeletal modulation and motion-guided temporal modulation to dynamically inject the motion patterns into spatial and temporal branches, respectively. Then, these features are aggregated to advance the spatio-temporal feature representation. 
Subsequently, we present a motion consistency learning module to capture motion cues from fine-to-coarse features. 
Finally, the learned features are used for micro-action classification.

\subsection{Feature Embedding}
\label{sec:embedding}

Given an input skeleton sequence, we uniformly sample it to a fixed temporal length $T$, resulting in a tensor $\mX_{raw} \in \R^{T \times V \times C_{in}}$, where $V$ denotes the number of joints and $C_{in}$ the input feature dimension. 
Considering the characteristics of subtle motion patterns of micro-actions, we present a set of data augmentation named Skeleton-Temporal Context-aware Augmentation (STCA). Specifically, STCA consists of skeleton-wise and temporal-wise augmentations. 
In the skeleton part, we apply random affine transformation (\ie, rotation, scaling, and translation) to each frame:
\begin{equation}
\mX_{sp,t} = \texttt{S}(\mX_{raw,t} R_\theta^T) + \epsilon,
\end{equation}
where $R_{\theta}$ is a rotation matrix, with the rotation angle $\theta \sim \mathcal{U}(-15^\circ, 15^\circ)$\sloppy. The scaling factor $S(\cdot) \sim \mathcal{U}(0.9, 1.1)$, and the translation offset $\boldsymbol{\epsilon} \sim \mathcal{U}(-0.1, 0.1)$. This operation is then broadcast to all joints, helping to mitigate spatial deviations caused by sensor noise or posture changes. 
In the temporal part, we jitter each frame in the sequence with small perturbations:
\begin{equation}
\mX_{aug,t} = \mX_{sp,\min(\max(t + \delta_t, 0), T - 1)},
\end{equation}
where $\delta_t$ is sampled from uniform distribution $\mathcal{U}(-3, 3)$. This operation enhances robustness to subtle timing shifts inherent in micro-motions. The augmented skeleton sequence is denoted as $\mX_{aug}\in\R^{T\times V\times C_{in}}$. Then, we utilize a projection module consisting of three linear layers to map the low-dimensional data into a higher-dimensional feature space $\mX_{proj}\in\R^{T\times V\times C}$.

To preserve the structural integrity of the skeleton data, we adopt Skate-Embedding~\cite{do2024skateformer} to build skeletal-temporal positional embeddings. 
Let the temporal indices be $T_{idx} = [t_1, t_2, \ldots, t_T]$, these indices are first normalized to the range $[-1, 1]$ and then processed using sinusoidal positional embeddings~\cite{vaswani2017attention} to generate fixed temporal features $\mF_{te} \in \R^{T \times C}$. 
Simultaneously, learnable skeletal features $\mF_{se} \in \R^{V \times C}$ are employed to encode the joint positional information, where the indices of the joints are used. 
The final skeletal-temporal positional embedding $\mF_{\mathrm{ste}} \in \R^{T \times V \times C}$ is computed by the outer product:
\begin{equation}
\mF_{\mathrm{ste}}[t, v, c] = \mF_{\mathrm{te}}[t, c] \cdot \mF_{\mathrm{se}}[v, c],
\end{equation}
where $t$, $v$, and $cs$ denote the temporal step, joint index, and channel index, respectively.

Finally, the skeletal-temporal positional embedding $\mF_{ste}$ is added to the projected skeleton features $\mX_{proj}$ to incorporate both spatial and temporal positional information. We denote the resulting input to the subsequent Motion-guided Feature Modulation as:
\begin{equation}
\mX_\mathrm{feat} = (\mX_\mathrm{proj} + \mF_\mathrm{ste}) \in \R^{T \times V \times C}.
\end{equation}

\subsection{Motion-guided Feature Modulation}
\label{sec:mfb}
To capture the subtle motion patterns in the skeleton sequence, we propose a Motion-guided Feature Modulation (MFM) module that models complex motion dynamics to guide the learning of compact spatio-temporal features.
As shown in Figure~\ref{fig:method}, MFM is composed of $N$ stacked Motion-guided Skeletal-Temporal Formers (MSTF). 
Inspired by Transformer-based methods~\cite{wang20233mformer,do2024skateformer} in skeleton-based action recognition, we utilize three parallel branches—namely, the \colorbox{MSMblue}{skeletal branch}, \colorbox{Mgreen}{motion branch}, and \colorbox{MTMyellow}{temporal branch}—to construct compact spatio-temporal representations. The input is first normalized and projected as:
\begin{equation}
\mX_{in} = \texttt{Linear}(\texttt{LN}(\mX_{feat})) \in \R^{T \times V \times C}.
\end{equation}

In the skeletal branch, we leverage a one-layer graph convolutional network (GCN) to model the relational structure among skeletal joints via a learnable adjacency matrix $\mA \in \mathbb{R}^{V \times V}$. This operation is formulated as:
\begin{equation}
\mX_{gc} = \texttt{GConv}(\mX_{in}[...,:C/4]) = \sigma(\mW_g (\mA \, \mX_{in}[...,:C/4])),
\label{eq:gconv}
\end{equation}
where $\mW_g$ denotes the learnable weight matrix, and $\sigma(\cdot)$ represents a nonlinear activation function (\ie, GELU~\cite{hendrycks2016gaussian}). Through the adjacency matrix, the $\texttt{GConv}$ branch aggregates information from adjacent joints, thereby extracting spatial features that characterize the skeletal structure. 

In the temporal branch, we apply a single-layer temporal convolution (T-Conv) along the temporal dimension of the sequence to capture local temporal dynamics, defined as:
\begin{equation}
\mX_{tc} = \texttt{TConv}(\mX_{in}[...,3C/4:]),
\label{eq:tconv}
\end{equation}
where $\texttt{TConv}$ aims to capture local temporal dynamics. This branch emphasizes short-range dependencies within the action sequence, enhancing the representation of micro-actions across the temporal dimension. 
The motion branch captures motion-specific features by computing the motion from consecutive frames. Specifically, the motion $\Delta \mX$ can be calculated by:
\begin{equation}
\Delta \mX = \mX_{in}[1:, :, C/4:3C/4] - \mX_{in}[:-1, :, C/4:3C/4].
\label{eq:delta_x}
\end{equation}

Subsequently, we utilize the motion-guided modulation network to decompose the motion information into skeletal and temporal branches. 

\textbf{Motion-guided Skeletal Modulation (MSM).} 
We use graph convolution to capture inter-joint relationships. 
Then, a Batch Normalization layer and $\tanh$ activation function are utilized to build modulation factors. This process can be formulated as: 
\begin{equation}
\mZ_s = \texttt{MSM}(\Delta \mX) = \tanh(\texttt{BN}(\texttt{GConv}(\Delta \mX))) \in \R^{T \times V \times C/2}, 
\end{equation}
\begin{equation}
\gamma_s, \beta_s = \mZ_s[..., :C/4],\ \mZ_s[..., C/4:], 
\end{equation}
where $\gamma_s$ and $\beta_s$ are scaling and shifting modulation factors, respectively. Then, we use these modulation factors to guide the temporal feature modeling:
\begin{equation}
\mX_{gcm} = \frac{\mX_{gc} - \mu(\mX_{gc})}{\sigma(\mX_{gc})} \cdot (1 + \gamma_{s}) + \beta_{s}.
\label{eq:modulation_t}
\end{equation}

\textbf{Motion-guided Temporal Modulation (MTM).}
This module employs a 2D convolution followed by a Batch Normalization (BN) layer and a $\tanh$ activation function to capture temporal dynamics. This process can be formulated as:
\begin{equation}
\mZ_t = \texttt{MTM}(\Delta \mX) = \tanh(\texttt{BN}(\texttt{Conv2d}(\Delta \mX))) \in \R^{T \times V \times C/2}, 
\end{equation}
\begin{equation}
\gamma_t, \beta_t = \mZ_t{[..., :C/4]},\ \mZ_t[..., C/4:], 
\end{equation}
where $\gamma_t$ and $\beta_t$ are scaling and shifting modulation factors, respectively. Then, we use these modulation factors to guide the temporal feature modeling:
\begin{equation}
\mX_{tcm} = \frac{\mX_{tc} - \mu(\mX_{tc})}{\sigma(\mX_{tc})} \cdot (1 + \gamma_{t}) + \beta_{t},
\label{eq:modulation_s}
\end{equation}
where $\mu(\mX_{tc})$ and $\sigma(\mX_{tc})$ are the mean and standard deviation of feature $\mX_{tc}$. 

Up to now, we have obtained the motion-guided skeletal feature $\mX_{gcm}$, motion-guided temporal feature $\mX_{tcm}$, and motion feature $\Delta \mX$. 
Considering that the motion feature contains essential temporal and spatial motion patterns, we adaptively aggregate these features to build the spatio-temporal skeleton features through a gating mechanism:
\begin{equation}
\mX_{agg} = \texttt{Aggregation}([\mX_{gcm}, \Delta \mX, \mX_{tcm}]) \in \R^{T \times V \times C},
\label{eq:aggregation}
\end{equation}
where $[\cdot]$ denotes channel-wise concatenation, $\texttt{Aggregation}(\cdot)$ learns different weights based on the feature distributions of each branch. This strategy helps balance contributions from spatial, temporal, and motion-specific features. 
To further enhance network stability, the fused features $\mX_{agg}$ pass through a projection layer and are then added via a residual connection to the original input $\mX_{raw}$. A FeedForward network~\cite{vaswani2017attention} is subsequently applied to introduce nonlinear transformations. 
Overall, the stacked $N$-layer Motion-guided Skeletal-Temporal Former preserves deeper motion cues, thereby strengthening the final representations $\mX_{f}$ of micro-action movements. 

\subsection{Motion Consistency Learning}\label{sec:mcl}
Micro-actions span a wide range of categories, specifically, 52 action-level categories derived from 7 body-level categories (\eg, head, hand, and body). As a result, different micro-actions exhibit variations in temporal duration. 
To capture motion patterns across multiple temporal scales, we construct $L$ spatio-temporal features ($\mX_{f}^{1}, \mX_{f}^{2}, \ldots, \mX_{f}^{L}$) by progressively downsampling the temporal dimension by a factor of two at each stage. 
Then, each scale feature is input to the Motion-guided Feature Modulation to build coarse-to-fine spatio-temporal features. 
Specifically, the unified temporal length is denoted as $T^\prime = T / 2^{(L-1)}$.
Subsequently, we utilize an adaptive gated fusion module to aggregate multi-scale representations:
\begin{equation}
\mX_{z} = \texttt{Aggregation}([\mX_{f}^{1}, \mX_{f}^{2}, \ldots, \mX_{f}^{L}]) \in \R^{T^\prime\times V\times C}.
\end{equation}

This adaptive fusion enables robust representation learning, integrating complementary information across temporal resolutions, thereby significantly improving the network’s sensitivity to subtle micro-action dynamics. Finally, the learned spatio-temporal features are fed into an action classifier to predict the final micro-action category.
Cross-entropy loss is utilized to optimize the model.

\textbf{Inference.} 
For each video, we input the corresponding skeleton data $\mX_{raw}$ to the proposed MMN and predict the action-level category. For the body-level category, we follow the practice in~\cite{guo2024benchmarking} to get the corresponding body-level results. 

%%%%%%%%%%%%%%%%%%%%%%%%%%%%%%%%%%%%%
\section{Experiments}
\subsection{Experimental Setup}
\textbf{Datasets.}
\textbf{Micro-Action 52}~\cite{guo2024benchmarking} is a large-scale whole-body micro-action dataset collected by a professional interview to capture unconscious human micro-action behaviors. The dataset contains 22,422 (22.4K) samples, where the annotations are categorized into two levels: 7 \textit{body-level} and 52 \textit{action-level} micro-action categories. There are 11,250, 5,586, and 5,586 instances in the training, validation, and test sets, respectively. 
In addition, we also evaluate the proposed method on a micro-action-related dataset, \ie, \textbf{iMiGUE}~\cite{liu2021imigue}, which mainly consists of micro-gestures of the upper body. 
iMiGUE was collected from post-competition athlete interviews and contains 32 micro-gesture categories. While conceptually similar to micro-actions, these micro-gestures are restricted to the upper limbs. 
We use the skeleton data released by the MiGA Competition~\cite{chen20242nd}.
The dataset contains 12,893, 777, and 4,562 samples for training, validation, and testing, respectively. 

\noindent \textbf{Evaluation Metrics.}
Following the standard practice~\cite{guo2024benchmarking,li2025prototypical}, we adopt Top-1/-5 accuracy and F1 score to evaluate the performance for this task. 
Herein, for the F1 metric, F1$_{macro}$ calculates the unweighted mean for each action category, independent of the sample size for the specific category, and F1$_{micro}$ treats all samples equally and is less influenced by any category with a predominant number of micro-actions. 
Considering micro-actions involving body parts and micro-action categories, these two F1 scores are evaluated on both ``$body$'' and ``$action$'' granularity. 
F1$_{mean}$ is used as a general evaluation metric, and its formula is as follows:
\begin{equation} 
\rm F1_{\emph{mean}} = \frac{\rm F1_{\emph{macro}}^{\emph{body}} \!+\! \rm F1_{\emph{micro}}^{\emph{body}} \!+\! \rm F1_{\emph{macro}}^{\emph{action}} \!+\! \rm F1_{\emph{micro}}^{\emph{action}}}{4}.
\label{eq:map}
\end{equation}
Following common practice in micro-gesture recognition~\cite{liu2021imigue,chen2023smg}, we report both top-1 and top-5 accuracies.

\noindent \textbf{Implementation Details.}
For each video, we extract the 44 key points via the AlphaPose~\cite{alphapose} model. 
The temporal length $T$ is set to 64. For model training, we adopt the AdamW optimizer~\cite{loshchilov2017decoupled} with a base learning rate of 1$\times$10$^{-4}$, a minimum learning rate of 1$\times$10$^{-6}$, weight decay of 0.1, and a batch size of 256. A cosine annealing learning rate scheduler with 3 cycles is employed throughout the training process. To stabilize the initial phase of training, we apply a linear warm-up strategy over the first 20 epochs, where the learning rate gradually increases from 1$\times$10$^{-7}$ to the base value. 

\begin{table*}[t!]
\centering
\tabcolsep 5pt
\renewcommand\arraystretch{0.9}
\caption{Quantitative comparison results on the Micro-Action 52~\cite{guo2024benchmarking} and iMiGUE~\cite{liu2021imigue} datasets, where the best values are indicated in \textbf{BOLD}.
$\mR$ denotes the RGB modality, $\mJ$ denotes the \textit{joint} modality, $\mB$ represents the \textit{bone} modality, and 2$\mS$ denotes ensemble results of joint and bone modalities. }
\resizebox{1.0\linewidth}{!}{
\begin{tabular}{|>{\raggedleft\arraybackslash}p{2.2cm}>{\raggedright\arraybackslash}p{1.0cm}||c|c|cc|cc|cc|c|cc|}
\hline
\thickhline
\rowcolor{mygray}
& & & \multicolumn{8}{c|}{Micro-Action 52} & \multicolumn{2}{c|}{iMiGUE} \\ \cline{4-13}
\rowcolor{mygray}
& & & \textit{Body} & \multicolumn{2}{c|}{\textit{Action}} & \multicolumn{2}{c|}{\textit{Body}} & \multicolumn{2}{c|}{\textit{Action}} & \textit{All} & \multicolumn{2}{c|}{\textit{Gesture}} \\ \cline{4-13}
\rowcolor{mygray}
\multicolumn{2}{|c||}{\multirow{-3}{*}{Method}} & \multirow{-3}{*}{M} & Top-1 & Top-1 & Top-5 & F1$_{macro}$ & F1$_{micro}$ & F1$_{macro}$ & F1$_{micro}$ & F1$_{mean}$ & Top-1 & Top-5\\ \hline
\rowcolor{rgbgray}
C3D~\cite{C3D}\!\!\!&\!\!\!\pub{ICCV 2015} & $\mR$ & 74.04 & 52.22 & 86.97 & 66.60 & 74.04 & 40.86 & 52.22 & 58.43 & 30.13 & 58.84 \\
\rowcolor{rgbgray}
I3D~\cite{I3D}\!\!\!&\!\!\!\pub{CVPR 2017} & $\mR$ & 78.16 & 57.07 & 88.67 & 71.56 & 78.16 & 39.84 & 57.07 & 61.66 & 35.08 & 62.70 \\
\rowcolor{rgbgray} TSN~\cite{wang2016temporal}\!\!\!&\!\!\!\pub{ECCV 2016} & $\mR$ & 59.22 & 34.46 & 73.34 & 52.50 & 59.22 & 28.52 & 34.46 & 43.67 & 51.31 & 84.42 \\
\rowcolor{rgbgray} TSM~\cite{lin2019tsm}\!\!\!&\!\!\!\pub{ICCV 2019} & $\mR$ & 77.64 & 56.75 & 87.47 & 70.98 & 77.64 & 40.19 & 56.75 & 61.39 & 56.23 & 88.39 \\
\rowcolor{rgbgray} SlowFast~\cite{feichtenhofer2019slowfast}\!\!\!&\!\!\!\pub{ICCV 2019} & $\mR$ & 77.18 & 59.60 & 88.54 & 70.61 & 77.18 & 44.96 & 59.60 & 63.09 & - & - \\
\rowcolor{rgbgray}
TIN~\cite{shao2020temporal}\!\!\!&\!\!\!\pub{AAAI 2020} & $\mR$ & 73.26 & 52.81 & 85.37 & 66.99 & 73.26 & 39.82 & 52.81 & 58.22 & - & - \\
\rowcolor{rgbgray}
TimesFormer~\cite{bertasius2021space}\!\!\!&\!\!\!\pub{ICML 2021} & $\mR$ & 69.17 & 40.67 & 82.67 & 61.90 & 69.17 & 34.38 & 40.67 & 51.53 & 50.66 & 84.76 \\
\rowcolor{rgbgray} Video SwinT~\cite{liu2022video}\!\!\!&\!\!\!\pub{CVPR 2022} & $\mR$ & 77.95 & 57.23 & 87.99 & 71.25 & 77.95 & 38.53 & 57.23 & 61.24 & 60.11 & 86.62 \\
\rowcolor{rgbgray}
Uniformer~\cite{li2022uniformer}\!\!\!&\!\!\!\pub{ICLR 2022}& $\mR$ & 79.03 & 58.89 & 87.29 & 71.80 & 79.03 & 48.01 & 58.89 & 64.43 & 59.91 & 90.30 \\ 
\rowcolor{rgbgray}
MANet~\cite{guo2024benchmarking}\!\!\!&\!\!\!\pub{TCSVT 2024} & $\mR$ & 78.95 & 61.33 & 88.83 & 72.87 & 78.95 & 49.22 & 61.33 & 65.59 & 58.08 & 84.09 \\
\rowcolor{rgbgray} PCAN~\cite{li2025prototypical}\!\!\!&\!\!\!\pub{AAAI 2025} & $\mJ$ & 76.99 & 56.51 & 87.13 & 71.33 & 76.99 & 42.28 & 56.51 & 61.78 & - & - \\ 
\rowcolor{rgbgray} PCAN~\cite{li2025prototypical}\!\!\!&\!\!\!\pub{AAAI 2025} & $\mR$ & 79.36 & 60.03 & 87.65 & 72.37 & 79.36 & 43.29 & 60.03 & 63.76 & - & - \\ 
\rowcolor{rgbgray} PCAN~\cite{li2025prototypical}\!\!\!&\!\!\!\pub{AAAI 2025} & $\mJ$+$\mR$ & \textbf{82.30} & \textbf{66.74} & \textbf{91.75} & \textbf{77.02} & \textbf{82.30} & \textbf{53.83} & \textbf{66.74} & \textbf{69.97} & - & - \\ 
\hline
\multirow{3}{*}{ST-GCN~\cite{yan2018spatial}}\!\!\!&\multirow{3}{*}{\!\!\!\pub{AAAI 2018}} & $\mJ$ & 72.99 & 55.46 & 83.51 & 64.16 & 72.99 & 39.21 & 55.46 & 57.96 & 52.00 & 85.62\\
& & $\mB$ & 73.77 & 55.93 & 83.26 & 65.18 & 73.77 & 41.10 & 55.93 & 59.00 & 46.58 & 85.23 \\
& & 2$\mS$ & 75.87 & 59.31 & 85.95 & 68.24 & 75.87 & 44.38 & 59.31 & 61.95 & 54.03 & 87.68 \\ \hline
%%%%
\multirow{3}{*}{AAGCN~\cite{shi2020skeleton}}\!\!\!&\multirow{3}{*}{\!\!\!\pub{TIP 2020}} 
& $\mJ$ & 72.00 & 53.15 & 79.57 & 63.42 & 72.00 & 36.69 & 53.15 & 56.32 & 53.95 & 85.73 \\
& & $\mB$ & 70.00 & 50.73 & 79.04 & 60.48 & 70.00 & 34.27 & 50.73 & 53.87 & 48.75 & 85.01 \\
& & 2$\mS$ & 74.13 & 56.96 & 84.37 & 65.88 & 74.13 & 41.36 & 56.96 & 59.58 & 56.18 & 87.64 \\ \hline
%%%%%
\multirow{3}{*}{MS-G3D~\cite{liu2020disentangling}}\!\!\!&\multirow{3}{*}{\!\!\!\pub{CVPR 2020}} 
& $\mJ$ & 70.14 & 50.04 & 79.48 & 61.60 & 70.14 & 35.45 & 50.04 & 54.31 & 55.28 & 85.01 \\
& & $\mB$ & 68.22 & 48.73 & 79.00 & 58.96 & 68.22 & 35.57 & 48.73 & 52.87 & 48.16 & 83.36 \\
& & 2$\mS$ & 71.21 & 52.70 & 82.33 & 63.16 & 71.21 & 38.78 & 52.70 & 56.46 & 55.81 & 87.09 \\ \hline
%%%%%
\multirow{3}{*}{CTR-GCN~\cite{chen2021channel}}\!\!\!&\multirow{3}{*}{\!\!\!\pub{ICCV 2021}} 
& $\mJ$ & 75.49 & 57.12 & 84.78 & 67.91 & 75.49 & 41.59 & 57.12 & 60.53 & 55.90 & 88.05 \\
& & $\mB$ & 71.84 & 52.54 & 81.53 & 62.84 & 71.84 & 35.89 & 52.54 & 55.78 & 48.11 & 84.48 \\
& & 2$\mS$ & 76.01 & 59.06 & 86.05 & 68.46 & 76.01 & 43.38 & 59.06 & 61.73 & 56.12 & 88.36 \\ \hline
%%%%%
\multirow{3}{*}{DGST-GCN~\cite{duan2022dg}}\!\!\!&\multirow{3}{*}{\!\!\!\pub{arXiv 2022}} 
& $\mJ$ & 73.31 & 53.40 & 82.80 & 64.32 & 73.31 & 32.62 & 53.40 & 55.91 & 56.66 & 89.32 \\
& & $\mB$ & 72.61 & 53.78 & 81.74 & 62.99 & 72.61 & 37.65 & 53.78 & 56.76 & 49.61 & 86.39 \\ 
& & 2$\mS$ & 74.65 & 57.43 & 85.30 & 65.65 & 74.65 & 40.81 & 57.43 & 59.63 & \ebestr{58.11} & \ebestr{89.48} \\ \hline
%%%%%
\multirow{3}{*}{ST-GCN++~\cite{duan2022pyskl}}\!\!\!&\multirow{3}{*}{\!\!\!\pub{ACM MM 2022}} & $\mJ$ & 68.03 & 45.85 & 74.72 & 57.66 & 68.03 & 30.59 & 45.85 & 50.53 & 51.53 & 84.41 \\
& & $\mB$ & 70.37 & 51.36 & 81.10 & 61.23 & 70.37 & 36.32 & 51.36 & 54.82 & 46.51 & 84.50 \\ 
& & 2$\mS$ & 72.04 & 53.78 & 82.04 & 62.95 & 72.04 & 37.52 & 53.78 & 56.57 & 53.88 & 87.22 \\ \hline
\multirow{3}{*}{HD-GCN~\cite{lee2023hierarchically}}\!\!\!&\multirow{3}{*}{\!\!\!\pub{ICCV 2023}} 
& $\mJ$ & 74.56 & 58.13 & 83.85 & 66.08 & 74.56 & 42.86 & 58.13 & 60.41 & 53.31 & 86.96 \\
& & $\mB$ & 73.33 & 56.73 & 84.94 & 65.11 & 73.33 & 42.26 & 56.73 & 59.36 & 45.73 & 82.27 \\
& & 2$\mS$ & 75.76 & 60.19 & 86.90 & 67.32 & 75.76 & 44.50 & 60.19 & 61.94 & 53.44 & 87.33 \\ \hline
% %%%%%
\multirow{3}{*}{Koopman~\cite{wang2023neural}}\!\!\!&\multirow{3}{*}{\!\!\!\pub{CVPR 2023}} 
& $\mJ$ & 74.51 & 58.79 & 85.05 & 66.29 & 74.51 & 43.80 & 58.79 & 60.85 & 51.78 & 86.10 \\
& & $\mB$ & 71.05 & 52.06 & 83.12 & 63.29 & 71.05 & 38.81 & 52.06 & 56.30 & 43.23 & 82.05 \\
& & 2$\mS$ & 75.04 & 59.70 & 86.79 & 66.48 & 75.04 & 44.57 & 59.70 & 61.45 & 52.08 & 86.41 \\ \hline
%%%%%
\multirow{3}{*}{FR-Head~\cite{Zhou_2023_CVPR}}\!\!\!&\multirow{3}{*}{\!\!\!\pub{CVPR 2023}} 
& $\mJ$ & 75.35 & 58.93 & 85.18 & 67.84 & 75.35 & 44.67 & 58.93 & 61.70 & 55.11 & 88.08 \\
& & $\mB$ & 72.72 & 56.07 & 84.84 & 63.98 & 72.72 & 42.11 & 56.07 & 58.72 & 47.81 & 85.25 \\
& & 2$\mS$ & \ebestr{76.35} & \ebestr{61.17} & 86.99 & \ebestr{68.88} & \ebestr{76.35} & \ebestr{47.43} & \ebestr{61.17} & \ebestr{63.46} & 55.26 & 88.71 \\ \hline
% %%%%%
\multirow{3}{*}{SkateFormer~\cite{do2024skateformer}}\!\!\!&\multirow{3}{*}{\!\!\!\pub{ECCV 2024}} 
& $\mJ$ & 75.03 & 58.92 & 82.69 & 67.14 & 75.03 & 45.10 & 58.92 & 61.55 & 47.90 & 85.31 \\
& & $\mB$ & 70.96 & 50.95 & 83.35 & 62.98 & 70.96 & 36.83 & 50.95 & 55.43 & 42.79 & 82.92 \\
& & 2$\mS$ & 75.67 & 59.76 & \ebestr{87.27} & 68.33 & 75.67 & 45.58 & 59.76 & 62.34 & 47.74 & 85.38 \\ \hline
%%%%%
& & $\mJ$ & 77.77 & 61.33 & 89.22 & 71.25 & 77.77 & 48.56 & 61.33 & 64.73 & 59.84 & 89.76 \\
& & $\mB$ & 75.47 & 57.50 & 86.54 & 66.82 & 75.47 & 40.49 & 57.50 & 60.07 & 54.56 & 88.10 \\
\multicolumn{2}{|c||}{\multirow{-3}{*}{\textbf{MMN (Ours)}}} & 2$\mS$ & \bestr{78.52} & \bestr{62.71} & \bestr{89.83} & \bestr{71.86} & \bestr{78.52} & \bestr{48.27} & \bestr{62.71} & \bestr{65.34} & \bestr{60.21} & \bestr{90.11} \\ \hline
%%%%%
\end{tabular}}
\label{tab:main}
\end{table*}

\subsection{Main Comparison}
We comprehensively compare the proposed MMN model with current state-of-the-art RGB-based and skeleton-based action recognition methods. 
The experimental results are reported in Table~\ref{tab:main}. 
On the \textbf{MA-52} dataset, under the single joint modality ($\mJ$), the proposed MMN achieves a Top-1 accuracy of 61.33\% in action-level classification, significantly outperforming early methods such as FR-Head and SkakeFormer, which typically achieve around 58\%. 
For the bone modality ($\mB$), although performance varies across competing approaches, MMN consistently maintains strong results, indicating that bone motion features also play a vital role in micro-action recognition. 
Notably, when fusing the joint and bone modalities (2$\mS$), MMN further improves the action-level Top-1 accuracy to 62.71\%, and achieves the highest F1$_{mean}$ of 65.34\%, approaching the best results reported by RGB-based methods. 

We further evaluate MMN on the \textbf{iMiGUE} dataset, which focuses on micro-gesture recognition. MMN achieves the best Top-1 and Top-5 performance (\ie, 60.21\% and 90.11\%), which validates the effectiveness of the proposed motion-guided modulation strategy. 
It not only outperforms many existing methods under single-modality settings but also shows strong baseline performance across different input types.
When joint and limb features are fused in the 2$\mS$ configuration, MMN further enhances its ability to capture subtle micro-gesture details, leading to improved overall recognition performance. 
As shown in Table~\ref{tab:efficiency_accuracy}, the proposed MMN achieves lower parameters (1.23M) and significantly reduced inference latency (7.15 ms) compared to existing methods.

\begin{table}[t!]
\tabcolsep 5pt
\renewcommand\arraystretch{0.9}
\caption{Efficiency and accuracy comparison on MA-52 with $\mJ$ modality data.}
\resizebox{1.0\linewidth}{!}{
\begin{tabular}{|r||ccc|c|}
\hline\thickhline
\rowcolor{mygray}
Method & Params. (M) & FLOPs (G) & Time (ms) & F1$_{mean}$\\
\hline
ST‑GCN~\cite{yan2018spatial}      & 3.10 & 16.32 & 52.2  & 57.96 \\
ST‑GCN++~\cite{duan2022pyskl}     & 3.10 & 3.48 & 11.1  & 50.53 \\
Koopman~\cite{wang2023neural}      & 5.38 & 8.76 & 17.9  & 60.85 \\
FR-Head~\cite{Zhou_2023_CVPR}      & 1.45 & 3.60 & 18.5  & 61.70 \\
HD-GCN~\cite{lee2023hierarchically}       & 1.66 & 3.44 & 72.8  & 60.41 \\
SkateFormer~\cite{do2024skateformer}  & 2.03 & 3.62 & 11.5  & 61.55 \\
\hline
\textbf{MMN (Ours)} & \textbf{1.23} & \textbf{1.48} & \textbf{7.15} & \textbf{64.73} \\
\hline
\end{tabular}}
\label{tab:efficiency_accuracy}
\end{table}

\begin{table}[t!]
\tabcolsep 8pt
\renewcommand\arraystretch{0.9}
\caption{Ablation study on motion-guided skeletal-temporal former. MSM denotes motion-guided skeletal modulation, while MTM represents motion-guided temporal modulation.}
\resizebox{1.0\linewidth}{!}{
\begin{tabular}{|c|cc||ccc|c|}
\hline\thickhline
\rowcolor{mygray}
& & & \textit{Body} & \multicolumn{2}{c|}{\textit{Action}} & \textit{All} \\
\rowcolor{mygray}
\multirow{-2}{*}{Exp.} & \multirow{-2}{*}{MSM} & \multirow{-2}{*}{MTM} & Top-1 & Top-1 & Top-5 & F1$_{mean}$ \\ \hline
A1 & \xmark & \xmark & 76.10 & 58.90 & 86.65 & 61.08 \\ 
A2 & \ding{51} & \xmark & 76.15 & 58.07 & 87.47 & 61.67 \\
A3 & \xmark & \ding{51} & 76.80 & 59.02 & 87.31 & 62.57 \\
A4 & \ding{51} & \ding{51} & \textbf{77.77} & \textbf{61.33} & \textbf{89.22} & \textbf{64.73} \\ \hline
\end{tabular}}
\label{tab:abl_archs1}
\end{table}

\subsection{Ablation Studies}
In this section, we present a series of ablation studies to evaluate the proposed MMN. Unless otherwise stated, all experiments utilize the joint modality as input on the Micro-Action 52 dataset. 

\noindent\textbf{Motion-guided skeletal-temporal former.} 
This module leverages motion patterns to guide the interaction between skeletal and temporal features, thereby capturing subtle motion cues in micro-actions more effectively. 
As shown in Table~\ref{tab:abl_archs1},
variant A3, which adopts only the Motion-guided Temporal Modulation (MTM) module, improves the Top-1 accuracy by approximately 0.70\% and F1$_{mean}$ by 1.49\% compared to the baseline model (A1), indicating that MTM effectively captures subtle temporal dynamics.
While Variant A2, which solely incorporates the Motion-guided Skeletal Modulation (MSM) module, also yields some performance improvement, the gain is relatively limited, suggesting that MTM plays a more crucial role in modeling temporal dynamics.
Variant A4 (Ours), which integrates both MSM and MTM modules, achieves the best performance, with noticeable improvements across all modalities (Body, Action, All), reaching an F1$_{\text{mean}}$ of 64.73\%.

\noindent\textbf{Strategy of motion-guided skeletal and temporal modulation.} 
As introduced in \S~\ref{sec:mfb}, we present a motion-guided skeletal and temporal modulation that leverages motion patterns to guide spatio-temporal feature modeling. We evaluate the design with several variants and report the ablation results in Table~\ref{tab:abl_archs2}. 
From variants B1 and B2, we observe that removing either the scale or shift factors leads to a significant performance drop, \eg, decreases of 2.31\% and 2.69\% in F1$_{mean}$, respectively. 
Additionally, as shown in B3 and B4, replacing the learnable modulation with simple element-wise addition, concatenation, or Hadamard product also degrades performance. 

\noindent\textbf{Skeletal-temporal context-aware augmentation.}
To evaluate the effectiveness of our Skeletal-Temporal Context-aware Augmentation module, we conduct ablation studies on skeletal-level and temporal-level augmentations separately, as shown in Table~\ref{tab:abl_archs4}. 
Variant C1 serves as the baseline without any data augmentation. 
Variant C2 applies only skeletal-level augmentation, resulting in an increase of F1$_{mean}$ to 63.20\%, indicating that perturbations at the skeletal level enhance the model’s robustness to local spatial deformations. 
Variant C3 adopts only temporal-level augmentation and also yields performance improvement, with F1$_{mean}$ reaching 61.61\%, demonstrating its effectiveness in simulating temporal scale variations. 
C4 (Ours) combines both skeletal and temporal augmentations, achieving the best performance across all metrics: Top-1 accuracy improves to 77.77\%, Top-5 accuracy reaches 89.22\%, and F1$_{mean}$ rises to 64.73\%. 
These results suggest that skeletal and temporal context-aware augmentations are complementary, effectively mitigating overfitting caused by limited dataset size and long-tail distributions, while enhancing the model’s sensitivity to fine-grained motion variations.

\begin{table}[t!]
\tabcolsep 8pt
\renewcommand\arraystretch{0.9}
\caption{Ablation study on strategy of motion-guided skeletal and temporal modulation.}
\resizebox{1.0\linewidth}{!}{
\begin{tabular}{|c|c||ccc|c|}
\hline\thickhline
\rowcolor{mygray}
& & \textit{Body} & \multicolumn{2}{c|}{\textit{Action}} & \textit{All} \\
\rowcolor{mygray}
\multirow{-2}{*}{Exp.} & \multirow{-2}{*}{Method} & Top-1 & Top-1 & Top-5 & F1$_{mean}$ \\ \hline
B1 & w/o scale factor $\gamma$ & 76.42 & 58.56 & 87.68 & 62.42 \\
B2 & w/o shift factor $\beta$ & 76.24 & 58.41 & 87.00 & 62.04 \\ \hline
B3 & Addition & 76.91 & 59.54 & 88.04 & 62.74 \\ 
B4 & Concatenation & 74.94 & 59.10 & 88.45 & 62.37  \\ 
B5 & Hadamard Product & 75.49 & 57.81 & 86.56 & 61.21 \\ \hline
B6 & \textbf{MMN (Ours)} & \textbf{77.77} & \textbf{61.33} & \textbf{89.22} & \textbf{64.73} \\ \hline
\end{tabular}}
\label{tab:abl_archs2}
\end{table}

\begin{table}[t!]
\tabcolsep 6pt
\renewcommand\arraystretch{0.9}
\caption{Ablation study on skeletal-temporal context-aware augmentation.}
\resizebox{1.0\linewidth}{!}{
\begin{tabular}{|c|cc||ccc|c|}
\hline\thickhline
\rowcolor{mygray}
& & & \textit{Body} & \multicolumn{2}{c|}{\textit{Action}} & \textit{All} \\
\rowcolor{mygray}
\multirow{-2}{*}{Exp.} & \multirow{-2}{*}{Skeletal} & \multirow{-2}{*}{Temporal} & Top-1 & Top-1 & Top-5 & F1$_{mean}$ \\ \hline
C1 & \xmark & \xmark & 74.56 & 58.79 & 84.21 & 61.43 \\
C2 & \ding{51} & \xmark & 75.93 & 59.88 & 86.26 & 63.20 \\
C3 & \xmark & \ding{51} & 74.02 & 58.37 & 85.87 & 61.61 \\ 
C4 & \ding{51} & \ding{51} & \textbf{77.77} & \textbf{61.33} & \textbf{89.22} & \textbf{64.73} \\ \hline
\end{tabular}}
\label{tab:abl_archs4}
\end{table}

\begin{figure*}[t!]
\centering
\includegraphics[width=1\linewidth]{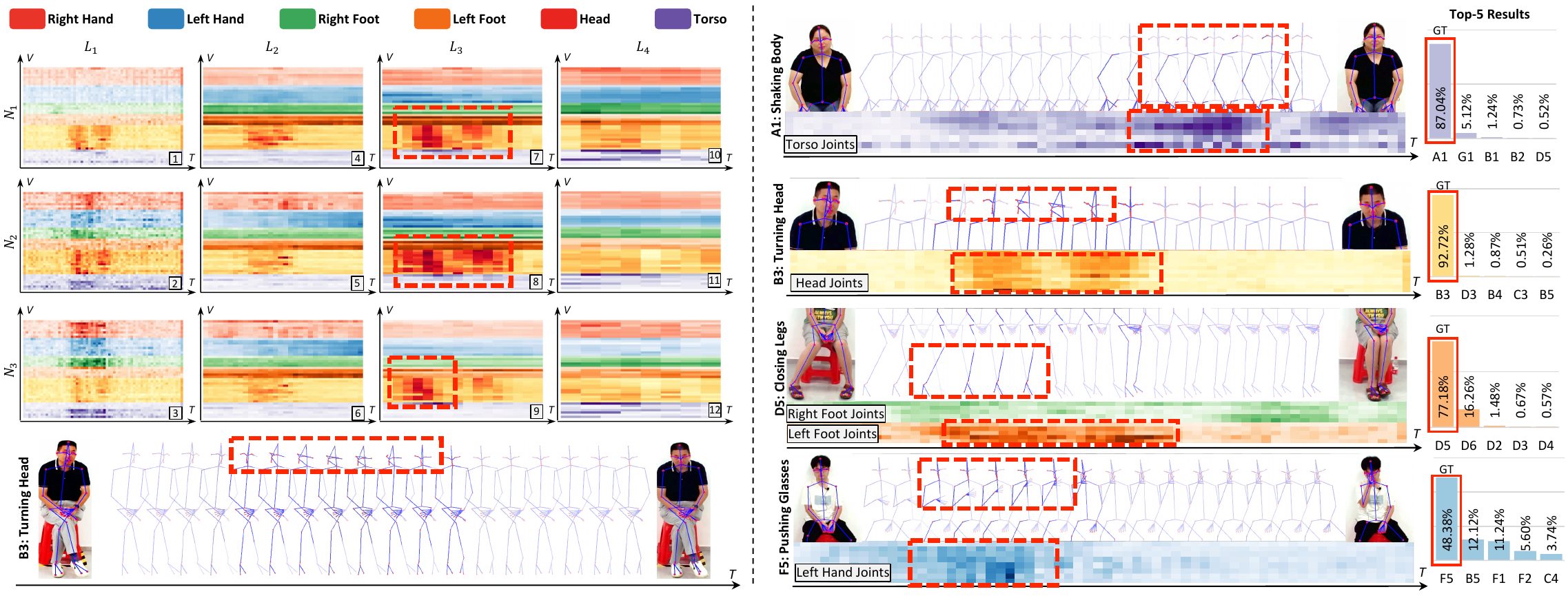}
\caption{Qualitative results on the Micro-Action 52 dataset~\cite{guo2024benchmarking}. 
\textbf{LEFT:} Motion-modulated feature $\mX_{agg}$ from $N$ stacked Motion-guided Skeletal-Temporal Formers (\S\ref{sec:mfb}). For the micro-action of ``turning head'' the model gradually focuses on discriminative joints, \ie, the facial joints highlighted by the red dashed box. \textbf{RIGHT:} Case study on different micro-actions.}
\label{fig:vis_results}
\end{figure*}

%%%%%%%%%%%%%%%%%%%%%%%%%%%%%%%%%%%%%%%%%%%%%%%%%%%%%%%%%%%%%%%%%%%%%%%%%%%%%
\subsection{Qualitative Analysis}
\noindent\textbf{The Visualization of Predicted Results.} 
In order to illustrate the effectiveness of the proposed Motion-guided Modulation Network in capturing subtle motion patterns, we give qualitative results in terms of motion-modulated spatio-temporal features in motion-guided feature modulation (\S\ref{sec:mfb}) and several case studies.

The left subfigure of Figure~\ref{fig:vis_results} shows motion-modulated feature maps $\tX_{agg}\in\R^{T\times V\times C}$ from each motion-guided skeletal-temporal former (MSTF). These features are aggregated by channel-wise max pooling. 
The embedding skeleton feature $\tX_{proj}$ is input to the stacked ($L_1 \rightarrow L_2 \rightarrow L_3 \rightarrow L_4$) motion-guided skeletal-temporal former for spatio-temporal feature modeling. In the motion-guided skeletal-temporal former, the feature is modulated sequentially ($N_1 \rightarrow N_2 \rightarrow N_3$).
We can see that the model progressively focuses on critical spatio-temporal joints relevant to the action. 
The ground truth label for this micro-action is ``B3: Turning head''. It can be observed that the facial joints (highlighted in yellow) are significantly emphasized in the 7th, 8th, and 9th feature maps. 
The interesting results indicate that the proposed method is able to capture fine-grained motion cues at both skeletal and temporal dimensions. 
Furthermore, we present additional case studies in the right subfigure of Figure~\ref{fig:vis_results}, covering the micro-actions from different body-level categories, \ie, ``A: body'', ``B: head'', ``D: Leg'', and ``F: Head-hand''. 
For each sample, we show the ground truth label, original frames, top-5 predicted classes, and the corresponding amplified action-related feature maps.  
We can see that the model can capture discriminative temporal segments and skeletal joints. 
Taking ``D5: Closing legs'' as an example, we observe that the legs gradually move closer together, with the left leg exhibiting a large motion. 
Correspondingly, the feature map highlights the left leg region, confirming the model’s sensitivity to localized motion variations. 
Overall, these qualitative results validate the effectiveness of the proposed method in capturing fine-grained motion patterns under the guidance of motion.

\begin{figure}[t!]
\centering
\includegraphics[width=0.60\linewidth]{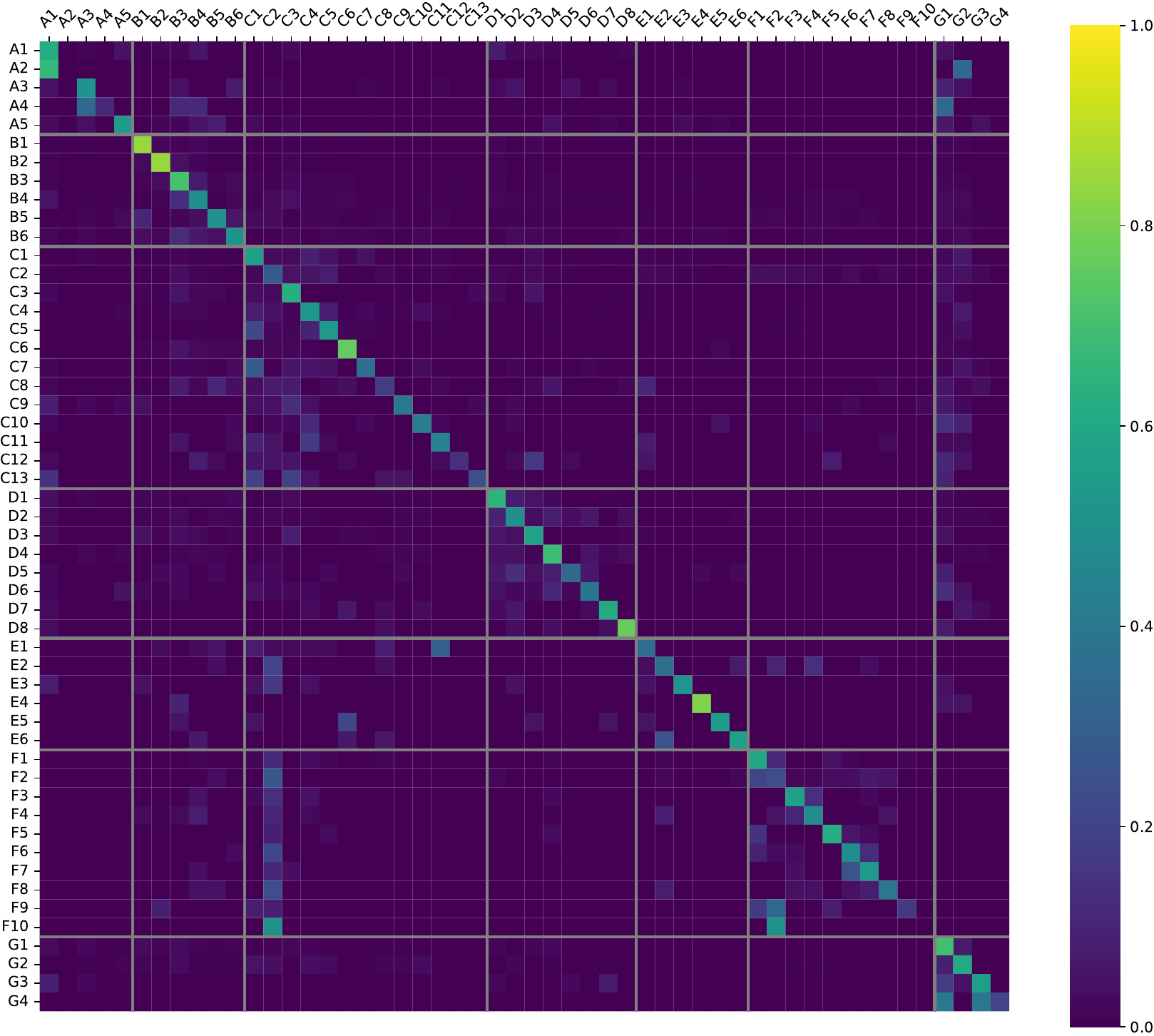}
\caption{Confusion Matrix of the test set on the Micro-Action 52 dataset. Please Zoom in for details.}
\label{fig:cm_visualization}
\vspace{-1.0em}
\end{figure}

\noindent\textbf{The confusion matrix of the predicted results.}
As shown in Figure~\ref{fig:cm_visualization}, we illustrate the confusion matrix of the action-level prediction results on the test set of the Micro-Action 52 dataset. Each matrix illustrates the distribution of predicted versus ground-truth classes, where the x-axis denotes predicted labels and the y-axis denotes ground-truth labels. Diagonal entries represent correct classifications, while off-diagonal entries indicate misclassifications. 
The results validate that the proposed method is effective in distinguishing fine-grained motion patterns and identifying inter-class actions across different body-level categories.

%%%%%%%%%%%%%%%%%%%%%%%%%%%%%%%%%%%%%
\section{Conclusion}
In this paper, we proposed a Motion-guided Modulation Network (MMN) specialized for Skeleton-based Micro-Action Recognition. To inject the motion for spatio-temporal feature representation, MMN decomposes motion information into two distinct levels: the skeletal level and the temporal level. At the skeletal level, the Motion-guided Skeletal Modulation (MSM) module reinforces the skeletal features of key joints, and the Motion-guided Temporal Modulation (MTM) module captures motion variations between consecutive frames. In addition, a motion consistency learning module is proposed to aggregate fine-grained motion patterns within multi-scale features.

% %% consistent spelling of the heading.
\begin{acks}
This work was supported by Anhui Provincial Key Research and Development Project (202304a05020068), the National Natural Science Foundation of China (62472381, U24A20331), the Postdoctoral Fellowship Program of CPSF (GZC20251086), the Fundamental Research Funds for the Central Universities of China (PA2025IISL0109), the Earth System Big Data Platform of the School of Earth Sciences, Zhejiang University, and the HPC Platform of Hefei University of Technology.
\end{acks}

\bibliographystyle{ACM-Reference-Format}
\balance
\bibliography{sample-base}

\end{document}

%% file: math_commands.tex
\usepackage{amsmath,amsfonts,bm}

\def\eqref#1{equation~\ref{#1}}
\def\1{\bm{1}}

\def\mA{{\bm{A}}}
\def\mB{{\bm{B}}}

\def\mF{{\bm{F}}}

\def\mJ{{\bm{J}}}

\def\mR{{\bm{R}}}
\def\mS{{\bm{S}}}

\def\mW{{\bm{W}}}
\def\mX{{\bm{X}}}

\def\mZ{{\bm{Z}}}

\DeclareMathAlphabet{\mathsfit}{\encodingdefault}{\sfdefault}{m}{sl}
\SetMathAlphabet{\mathsfit}{bold}{\encodingdefault}{\sfdefault}{bx}{n}
\newcommand{\tens}[1]{\bm{\mathsfit{#1}}}

\def\tX{{\tens{X}}}

\def\gM{{\mathcal{M}}}

\newcommand{\R}{\mathbb{R}}